\documentclass[runningheads]{llncs}
\usepackage[T1]{fontenc}
\usepackage{comment}
\usepackage{amsmath}
\usepackage{hyperref}
\usepackage{appendix}
\usepackage{multirow}
\usepackage{algorithm}
\usepackage{algorithmic}
\usepackage{graphicx}
\usepackage{float}
\newcommand{\myeq}[1]{\hfill{\refstepcounter{equation}(\theequation)\label{#1}}}

\begin{document}

\title{FineFACE: Fair Facial Attribute Classification Leveraging Fine-grained Features\thanks{Supported by National Science Foundation, USA}}

\titlerunning{Fair Facial Attribute Classification using Fine-grained Features}

\author{Ayesha Manzoor \and Ajita Rattani\thanks{Corresponding author.}}

\authorrunning{Ayesha Manzoor and Ajita Rattani}

\institute{Dept. of Computer Science and Engineering, \\ University of North Texas at Denton, Texas, USA \\ \email{\{ayeshamanzoor\}@my.unt.edu};~\email{\{ajita.rattani\}@unt.edu}}
\maketitle
\begin{abstract}
Published research highlights the presence of demographic bias in automated facial attribute classification algorithms, particularly impacting women and individuals with darker skin tones. Existing bias mitigation techniques typically require demographic annotations and often obtain a trade-off between fairness and accuracy, i.e., Pareto inefficiency. 
Facial attributes, whether common ones like gender or others such as "chubby" or "high cheekbones", exhibit high interclass similarity and intraclass variation across demographics leading to unequal accuracy. This requires the use of local and subtle cues using fine-grained analysis for differentiation. 
This paper proposes a novel approach to fair facial attribute classification by framing it as a fine-grained classification problem. Our approach effectively integrates both low-level local features (like edges and color) and high-level semantic features (like shapes and structures) through cross-layer mutual attention learning. Here, shallow to deep CNN layers function as experts, offering category predictions and attention regions. An exhaustive evaluation on facial attribute annotated datasets demonstrates that our FineFACE model improves accuracy by $1.32\%$ to $1.74\%$ and fairness by $67\%$ to $83.6\%$, over the SOTA bias mitigation techniques. Importantly, our approach obtains a Pareto-efficient balance between accuracy and fairness between demographic groups. In addition, our approach does not require demographic annotations and is applicable to diverse downstream classification tasks. To facilitate reproducibility, the code and dataset information is available at~\url{https://github.com/VCBSL-Fairness/FineFACE}.

\end{abstract}
\begin{figure}
    \centering
    \includegraphics[width=0.4\linewidth]{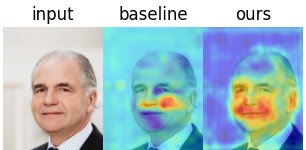}
    \includegraphics[width=0.4\linewidth]{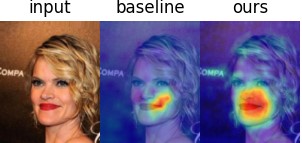}
    \includegraphics[width=0.4\linewidth]{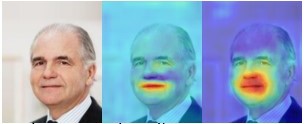}
    \includegraphics[width=0.4\linewidth]{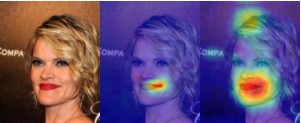}
    \caption{Visualization of the attention map obtained by our proposed FineFACE over baseline (both using ResNet50 backbone) for facial attribute classification. The highly activated region is shown by the red zone on the map, followed by yellow, green, and blue zones. Top: "High Cheekbones" classifier. Bottom: "Smiling" classifier.}
    \label{Figure:my_cam_viz}
    \vspace{-15pt}
\end{figure}

\section{Introduction}

Automated facial analysis algorithms encompass face detection, face recognition, and facial attribute classification (such as gender, race, high cheekbones, and attractiveness)~\cite{facerecsurvey,facedetectionsurvey,liu2015faceattributes}. These algorithms are deeply integrated into various sectors, such as surveillance and border control, retail and entertainment, healthcare, and education.

Numerous existing studies~\cite{AlbieroSVZKB20,KrishnanAR20,ssl-ijcb} investigating the \textit{fairness of facial attribute classification} algorithms confirm the presence of \textit{performance disparities between demographic groups, such as gender and race}. 
Thus, bias in these algorithms emerges as a significant societal issue that warrants immediate redress, particularly for the large-scale deployment of fair and trustworthy systems across demographics.
In this direction, the vision community has proposed several bias mitigation techniques to address the performance disparities of facial attribute classifiers.  
Established bias mitigation techniques utilize regularization~\cite{KrishnanR23}, attention mechanism~\cite{MajumdarSV21}, adversarial debiasing~\cite{ZhangLM18,chuang2021fair}, GAN-based over-sampling ~\cite{zietlow2022leveling,ramachandran2022deep}, multi-task classification~\cite{DasDB18}, and network pruning~\cite{LinKJ22}. 

These existing bias mitigation techniques often require demographically annotated training sets and are limited in their generalizability. Importantly, these techniques often sacrifice overall classification accuracy in pursuit of improved fairness, making them \textit{Pareto inefficient} \cite{zietlow2022leveling,ZhangLM18}. It was demonstrated in~\cite{zietlow2022leveling} that fairness violations in vision models are largely driven by the variance component of bias-variance decomposition. Consequently, \textit{one effective way to improve fairness is by decreasing the variance within each demographic subgroup by focusing on local and subtle cues}. This can be obtained through learning enhanced \textbf{feature representation} for each demographic subgroup, also supported by~\cite{cui2024classes,KrishnanAR20}. 

Following this line of thought, enhancing feature representation for each demographic subgroup is crucial in improving fairness without compromising overall performance. Traditional facial attribute classifiers~\cite{gendershade,KrishnanAR20,Muthukumar19,chuang2021fair} rely predominantly on high-level discriminative and semantically meaningful information often obtained from the final layers of the deep convolutional neural network (CNN).  However, the lower layers of the deep learning model capture (low-level) essential features and patterns in faces vital for attribute classification, such as \textbf{(a)} facial contours and edges, including the outline of the face, jawline and cheekbones, \textbf{(b)} texture of facial regions, such as skin and hair, \textbf{(c)} position and shape information, and \textbf{(d)} lighting condition and its effect on the appearance of facial features. Integrating low-level details from the lower layers of the model will capture local and detailed cues in the learned feature representation.

In our quest to identify these subtle and local cues for learning enhanced feature representation, we \textbf{aim} to leverage fine-grained analysis, integrating both high- and low-level features, toward fair facial attribute classification, FineFACE. This is facilitated through a cross-layer mutual attention learning technique that learns fine-grained features by considering the layers of a deep learning model from shallow to deep as independent 'experts' knowledgeable about low-level detailed to high-level semantic information, respectively. These experts are trained in leveraging mutual data augmentation to incorporate attention regions proposed by other experts. An ordinary deep learning model can be considered to use only the deepest expert (using high-level semantic information) for classification. In contrast, our method consolidates the prediction of the categorical label and the attention region of each expert for the final facial attribute classification task.
  
  Fig.~\ref{Figure:my_cam_viz} shows the final CAM visualization obtained by our proposed FineFACE model based on the ResNet50 backbone, for facial attribute classification with "high cheekbone" and "smiling" as target variable using the CelebA dataset~\cite{liu2015faceattributes}. The highly activated region is shown by the red zone on the map, followed by yellow, green, and blue zones in the attention map.  
   For cross-comparison, the visualization of the baseline ResNet$50$ is also shown for the same classification task. As illustrated in the maps, our FineFace model captures additional information, such as the contours of facial regions derived from the lower layers of the model, leading to enhanced feature representation and, hence, accurate and fair facial attribute classification.

\noindent \textbf{Contributions.} In summary, the contributions of our work are as follows: \textbf{(i)} We approach fair facial attribute classification from a novel perspective by reformulating it as a fine-grained classification task, \textbf{(ii)} We propose a novel approach based on cross-layer mutual attention learning where the prediction is consolidated from shallow (using low-level details) to deep layers (using high-level semantic details) regarded as an independent experts, 
\textbf{(iii)} Extensive evaluation on facial attribute annotated datasets namely, FairFace~\cite{karkkainen2021fairface}, UTKFace~\cite{Zhang_2017_CVPR}, LFWA+~\cite{liu2015faceattributes}, and CelebA~\cite{liu2018large}, 
and \textbf{(iv)} Cross-comparison with the existing bias mitigation techniques, demonstrating the efficacy of our approach in terms of significant improvement in fairness as well as classification accuracy.  Thus, obtaining state-of-the-art Pareto-efficient performance.

\section{Related Work}
In this section, we review the related academic literature.\\

\noindent \textbf{Bias Mitigation of Facial Attribute Classification.} Many studies have highlighted the systematic limitations of facial attribute classification (such as gender, race, and age) between gender-racial groups~\cite{gendershade,KrishnanAR20,Muthukumar19}. Studies in~\cite{chuang2021fair,RamaswamyKR21} reported bias of facial attribute classification for attractive, smiling, and wavy hair as the target attributes across genders. Following this study, ~\cite{zietlow2022leveling} reported bias of gender-independent target attributes, such as black hair, smiling, slightly open mouth, and eyeglasses, between genders.
Consequently, numerous strategies have been proposed to mitigate bias~\cite{bias_survey}.
~\cite{DasDB18} explored the joint classification of gender, age, and
race by proposing a multi-task network.~\cite{ZhangLM18} included a variable for the group of interest and simultaneously learned a predictor and an adversary via adversarial debiasing. 
~\cite{KrishnanR23} leveraged the power of semantic preserving augmentations at the image level in a self-consistency setting for fair gender classification tasks.~\cite{padala2020fnnc} introduced a framework that integrates fairness constraints directly into the loss function using Lagrangian multipliers for fair classification.
~\cite{chuang2021fair} proposed "fair mixup," a data augmentation technique by interpolating data points that improve the generalization of the classifiers trained under group fairness constraints.
~\cite{ramachandran2022deep} adopted structured learning techniques using deep-views of the training samples generated using GAN-based latent code editing to improve the fairness of the gender classifier.
GAN-based SMOTE "g-SMOTE" was proposed by~\cite{zietlow2022leveling} to strategically enhance the training set for underrepresented subgroups to mitigate bias.\\

\noindent \textbf{Fine-grained Visual Classification.}
Fine-grained classification is a challenging research task in computer vision, which captures the local discriminative features using attention learning~\cite{zhou2016learning,fu2017look} to distinguish different fine-grained categories. In addition to methods based on attention mechanisms, second-order pooling methods utilize the second-order statistics of deep features to compose powerful representations such as combined feature maps~\cite{lin2017bilinear} and covariance among deep features~\cite{wang2020deep} for fine-grained classification. 
Studies have also been proposed to use features or information learned from different layers within a CNN backbone for fine-grained classification. ~\cite{zeiler2014visualizing} proposed a multi-layered Deconvnet for gaining insight into the functions of intermediate feature layers. They discovered that shallow layers capture low-level details, whereas deep layers capture high-level information. ~\cite{jiang2021layercam} proposed the LayerCAM, which indicates the discriminative regions used by the different layers of a CNN to predict a specific category. Inspired by these two works,~\cite{liu2023learn} proposed the CMAL-Net, which focuses on using attention regions predicted by layers of different depths to mark the cues they learned, letting layers of varying depths to learn from each other's knowledge to improve overall performance.

\begin{figure}
    \centering
    \includegraphics[width=1\linewidth]{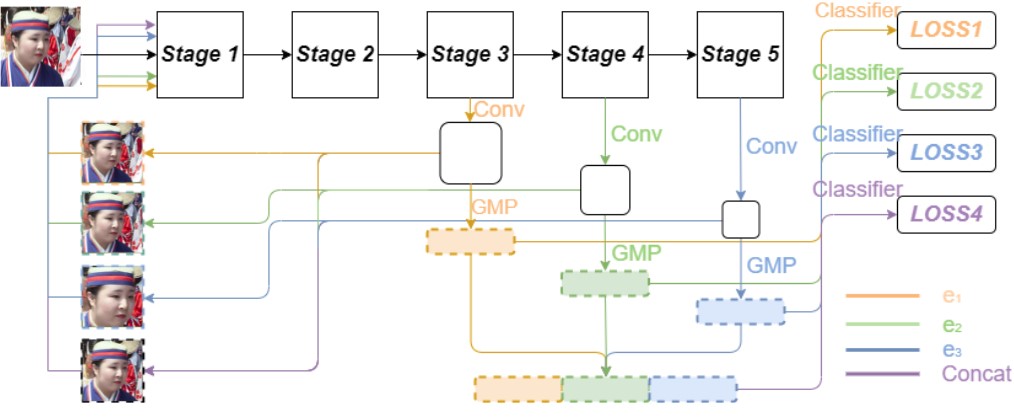}
    \caption{FineFACE network structure. This figure illustrates this method by introducing three experts $e_{1}$, $e_{2}$, $e_{3}$, on a 5-stage backbone CNN (e.g., ResNet50). The working of each expert and the concatenation of experts are depicted in different colors. Each expert receives feature maps from a specific layer as input and generates a categorical prediction along with an attention region, which is used for data augmentation by other experts. This architecture is trained in multiple steps within each iteration. We start by training the deepest expert (e3), followed by the shallower experts. Finally, in the last step, we train the concatenation of experts to enhance overall performance.}
    \label{fig:my_architecture}
%\vspace{-35pt}
\end{figure}

\section{Proposed Methodology}\label{secpm}

In this section, we elaborate on our proposed FineFACE model based on learning features from different layers of the CNN using the attention mechanism and mutual learning, following the foundational works in~\cite{zeiler2014visualizing,jiang2021layercam,liu2023learn}.

\subsection{Expert Construction: Using Shallow to Deep Layers}
In this subsection, the construction of experts from shallow to deep layers. Any state-of-the-art CNNs, such as ResNet50, Res2NeXt50, etc. can be used as the backbone CNN denoted by $\beta$. $\beta$  has $M$ layers, and \{$l_{1}$, $l_{2}$, ..., $l_{m}$, ..., $l_{M}$\} denote the layers
of $\beta$ from shallow to deep (except the fully connected layers).
\{$e_{1}$, $e_{2}$, ..., $e_{n}$, ..., $e_{N}$\} are N experts based on these M layers. Each expert encompasses layers from the first layer up to a certain layer such that - $e_{n}$ consists of the layers from $l_{1}$ to $l_{m_{n}}$, and 1 $\leq$  $m_{n}$  $\leq$ M. The experts \{$e_{1}$, $e_{2}$, ..., $e_{n}$, ..., $e_{N}$\} progressively cover deeper layers of the backbone CNN, and $e_{N}$, the deepest expert, covers all layers from $l_{1}$ to $l_{M}$.

Let \{$x_{1}$, $x_{2}$, ..., $x_{n}$, ..., $x_{N}$\} denote the intermediate feature maps produced by $\beta$ for the experts \{$e_{1}$, $e_{2}$, ..., $e_{n}$, ..., $e_{N}$\}, respectively. $x_{n} \in {R}^{H_{n}\times W_{n}\times C_{n}}$ and $H_{n}$, $W_{n}$ and $C_{n}$ denote the height, width, and number of channels, respectively. A set of functions \{$F_{1}$(.), $F_{2}$(.), ..., $F_{n}$(.), ..., $F_{N}$(.)\} are used to respectively compress \{$x_{1}$, $x_{2}$,..., $x_{n}$, ..., $x_{N}$\} into 1D vectorial descriptors \{$v_{1}$, $v_{2}$, ..., $v_{n}$, ..., $v_{N}$\}, and $v_{n}$ $\in$  $R^{C_v}$. $C_{v}$ denotes the length of the 1D vectorial descriptors, and these descriptors given by various experts are of the same length. The $F_{n}(.)$ for processing $x_{n}$ is defined as:

$v_{n}$ = $F_{n}$($x_{n}$) = $f^{GMP}$($x_{n}^{''}$), \myeq{1}\\

$x_{n}^{''}$ = $f^{Elu}$($f^{bn}$($f_{3\times3\times C_{v/2}\times C_{v}}^{conv} (x_{n}^{'})))$, \myeq{2}\\

$x_{n}^{'}$ = $f^{Elu}$($f^{bn}$($f_{1\times1\times C_{n}\times C_{v/2}}^{conv} (x_{n})))$, \myeq{3}\\

where $f^{GMP}$(.) denotes the Global Max Pooling. $f^{conv}$(.) depicts the 2D convolution operation by its kernel size.
$f^{bn}$(.) and $f^{Elu}$(.) denote batch normalization and Elu operations respectively. $x_{n}^{'}$ and $x_{n}^{''}$ are intermediate feature maps produced by $e_{n}$. Thereafter, $x_{n}^{''} \in {R}^{H_{n}\times W_{n}\times C_{n}}$ is used to generate the attention region of $e_{n}$ as described in subsection \ref{subsecarp}.
\{$p_{1}$, $p_{2}$, ..., $p_{n}$, ..., $p_{N}$\} denote the prediction scores given by different experts, obtained as $p_{n}$ = $f_{n}^{clf}(v_{n})$, where $f_{n}^{clf}(.)$ denotes a fully connected layer-based classifier.

Apart from the prediction scores obtained by the various experts, an overall prediction score is also generated by combining the information from different experts. Specifically, \{$v_{1}$, $v_{2}$, ..., $v_{n}$, ..., $v_{N}$\} are first concatenated for an overall descriptor $v_{oval}$ as: $v_{oval}$ = $f_{concat}$($v_{1}$, $v_{2}$, ..., $v_{n}$, ..., $v_{N}$), where $f_{concat}$(.) denotes concatenation operation. Then $v_{oval}$ is processed into an overall prediction score $p_{oval}$ by a fully connected layer-based classifier as $p_{oval}$ = $f_{oval}^{clf}(v_{oval})$

\vspace{-4pt}
\subsection{Attention Region Prediction}\label{subsecarp}
As mentioned above, $x_{n}^{''}$ denotes an intermediate feature map generated by the expert $e_{n}$. We move with an assumption that the classification problem is $K$-class, and $k_{n}$ $\in$ {1, 2, ..., K} is the category predicted by expert $e_{n}$ and $x_{n}^{''}\in{R}^{H_{n}\times W_{n}\times C_{n}}$. The generation of the attention region proposed by $e_{n}$ is initialized by producing the class activation map (CAM), which specifies the discriminative image region, for the category $k_{n}$ based on $x_{n}^{''}$ specified.
The CAM $\Omega_{n}$ ($\Omega_{n} \in {R}^{H_{n}\times W_{n}}$) produced by the expert $e_{n}$ is defined as: $\Omega_{n}$($\alpha,\beta$) = $\sum\limits_{c=1}^{c_{v}} p_{n}^{c}x_{n}^{''c}(\alpha, \beta)$, 
where the coordinates ($\alpha$, $\beta$) denote the spatial location of $x_{n}^{''}$ and $\Omega_{n}$.$p_{n}$ denotes the parameters of $f_{n}^{clf}(.)$ corresponding to the predicted category $k_{n}$. 
Then, after obtaining $\Omega$, an attention map $\tilde{\Omega}_{n}$ $\in {R}^{H_{in}\times W_{in}}$ ($H_{in}$, $W_{in}$ are the height and width of the input image, respectively) is generated by upsampling $\Omega_{n}$ using a bilinear sampling kernel. Thereafter, $\tilde{\Omega}_{n}$ is applied with min-max normalization, and each spatial element of the normalized attention map $\tilde{\Omega}_{n}^{norm}$ is obtained by 

$\tilde{\Omega}_{n}^{norm}$($\alpha$, $\beta$) = {$\tilde{\Omega}_{n}$($\alpha$, $\beta$) - min($\tilde{\Omega}_{n}$)}/{max($\tilde{\Omega}_{n}$) - min($\tilde{\Omega}_{n}$)}. \myeq{5}\\

\noindent The regions that the expert $e_{n}$ considers discriminative can be found and cropped by generating a mask $\tilde{\Omega}_{n}^{mask}$ by setting the elements in $\tilde{\Omega}_{n}^{norm}$ to $1$ for values greater than a threshold $t$ ($t$ $\in$ [0, 1]) and $0$ for the others.

Then, a box that covers all the positive regions of $\tilde{\Omega}_{n}^{mask}$ is located and cropped from the input image. The cropped region is upsampled to the input image’s size and the upsampled attention region $A_{n}$ is considered as the attention region predicted by $e_{n}$ and also as data augmentation for remaining experts. Apart from the attention regions proposed by various experts, an overall attention region $A_{oval}$ is generated by summing up the attention information learned by different experts. 

\subsection{Multi-step Mutual Learning}
The experts are trained using progressive multi-step strategy with cross-entropy loss. %following~\cite{liu2023learn}. 
In the early steps, these experts are trained one by one, which allows them to “focus on” learning the clues of their own expertise without being diverted by other experts. In the last two steps, the experts get together to learn impactful information from the attention regions and the raw image, respectively. Specifically, every iteration of the training takes place in $N+2$ steps, and in the first $N$ steps, each expert is gradually trained from deep to shallow. In the first step, the deepest expert $e_{N}$ is trained. Since the training of $N$ involves the experts shallower than $e_{N}$, the attention regions proposed by all the experts and the overall attention region \{$A_{1}$, $A_{2}$, ..., $A_{n}$, ..., $A_{N}$, $A_{oval}$\} are also generated at this step. These attention regions showcase the "specialized knowledge" of the experts by highlighting the basis on which each expert made its classification.

From step $2$ to $N$, there is a progression to shallower experts by
randomly selecting one input from a pool of images comprising of the raw input and the attention regions proposed by the other experts.

The shallow experts rely on the attention regions proposed by deeper experts to learn semantic visual clues (e.g., eyes, nose, and mouth), while the deep experts take the help of shallow experts by learning low-level visual cues (e.g., facial contours like jawline, cheekbones, etc.) from their proposed attention regions.

In step $N+1$, all the experts and their concatenation are trained with the overall attention region $A_{oval}$ in one pass. This step enforces all experts to work together and study the attention information they have combinedly gained for learning more fine-grained features.  At step $N+2$, the concatenation of all the experts is trained with the raw input to make sure the parameters of $f_{oval}^{clf}(.)$  fit the resolution of the original input. The \textbf{algorithm} for the multi-step mutual learning strategy is included in Section ~\ref{sec3} of the supplementary material.

\noindent \textbf{Inference Phase:} Fig.~\ref{fig:my_architecture} illustrates the inference stage of the proposed FineFACE model with $N+1$ classifiers. For an input image during the inference, $N+1$ prediction scores are produced by the proposed architecture. For each test image, the raw input and overall attention region are successively fed to the model obtaining $2\times(N + 1)$ number of prediction scores. The final prediction score for the inference is the average of the $2\times(N + 1)$ scores. This inference strategy maximizes the classification accuracy as well as fairness of the trained model due to obtaining two kinds of complementary information: (a) information from the prediction scores from various experts and the overall prediction score, and (b) the information from the raw input and overall attention region. 

\section{Experimental Details}\label{secexpdeets}

 We conducted two sets of \textbf{experiments} \textbf{(1)} a face-based gender classifier with gender as the target attribute and race and gender as the protected attributes following studies in~\cite{KrishnanR23,ramachandran2022deep}.
\textbf{(2)} $13$ gender-independent facial attribute classifiers following studies in~\cite{zietlow2022leveling,RamaswamyKR21,chuang2021fair} with “bags under eyes”, “bangs”, “black hair”, “blond hair”, “brown hair”, “chubby”, “eyeglasses”, “gray hair”, “high cheekbones”, “mouth slightly open”, “narrow eyes”, “smiling”, and “wearing hat” as the $13$ gender independent target attributes and gender as the protected attribute. We used the mean scores of these $13$ attribute classifiers, following studies in~\cite{zietlow2022leveling,RamaswamyKR21,chuang2021fair}.
\vspace{-1pt}
\subsection{Datasets and Training Protocol} 

We used standard benchmark datasets widely adopted for evaluating fairness of facial attribute classifiers~\cite{KrishnanR23,zietlow2022leveling,RamaswamyKR21}, namely, FairFace~\cite{karkkainen2021fairface}, UTKFace~\cite{Zhang_2017_CVPR}, LFWA+~\cite{liu2015faceattributes}, and CelebA~\cite{liu2018large}.
In line with existing studies~\cite{KrishnanR23,ramachandran2022deep}, a face-based gender classifier was trained on FairFace and evaluated on FairFace, UTKFace, LFWA+, and CelebA ($40$ attributes). Unlike UTKFace and LFWA+, CelebA does not have race annotations. Hence, we used only the gender attribute for CelebA.
For the $13$ gender-independent facial attribute classifiers, we used the CelebA ($40$ attributes) dataset for training and validation. For the fair comparison with the existing studies on fairness~\cite{zietlow2022leveling,RamaswamyKR21,chuang2021fair}, we used only $13$ gender-independent attributes from CelebA. Note that protected attribute annotation information is not used during the model training stage, but solely for the purpose of fairness evaluation of the facial attribute classifiers. Additional details on these datasets are given in Table \ref{tab:dataset_details}
\begin{table}[ht]
\centering
\begin{tabular}{ccc}
\hline
\textbf{Dataset} & \textbf{Images} & \textbf{Demographic groups}                                                                                              \\ \hline
FairFace         & 108K            & \begin{tabular}[c]{@{}c@{}}White, Black, Indian, Asian, \\ Southeast Asian, Middle Eastern, Latino Hispanic\end{tabular} \\ \hline
UTKFace          & 20K             & White, Black, Indian, Asian, Others                                                                                      \\ \hline
LFWA+            & 13K             & White, Black, Indian, Asian, Undefined                                                                                   \\ \hline
CelebA           & 202K            & Not Available (only gender information available)                                                                        \\ \hline
\end{tabular}
\caption{Dataset details including the number of images and demographic groups}
\label{tab:dataset_details}
\end{table}

\vspace{-25pt}
\subsection{Implementations Details}
For a fair comparison with studies in~\cite{zietlow2022leveling,KrishnanAR20,ramachandran2022deep}, we utilized ResNet50~\cite{he2016deep} as our method's backbone CNN architecture. The layers of ResNet$50$, excluding the fully connected layers, are grouped into $5$ stages where each stage is a group of layers operating on feature maps of the same spatial size. We use these stages as building blocks for our experts: $e_{1}$ encompasses layers from stage $1$ to stage $3$, $e_{2}$ encompasses layers from stage $1$ to stage $4$, and $e_{3}$ encompasses layers from stage $1$ to stage $5$. In general, the number of stages in a model can be determined by grouping layers operating on the feature maps of the same spatial size, and accordingly, experts can be formed.

We trained all the models used in this study using Stochastic Gradient Descent (SGD) with the number of epochs determined using an early stopping mechanism, the momentum of $0.9$, weight decay of $5\times10^{-4}$, and a mini-batch size of $16$ determined using empirical evidence. The learning rate was set as $0.002$ with cosine annealing~\cite{loshchilovstochastic}. We fixed the input image size as $448\times448$, following the common settings in existing fairness studies~\cite{gao2020channel,luo2020learning}. The threshold $t$, which is used to generate a mask for the attention region, was set to $0.5$ (see Section~\ref{subsecarp}). We also conducted an \textit{ablation study} of the related design choices (a) different pooling methods for building experts, and (b) the contribution of fusing the prediction scores. 
See~\textbf{ablation studies} in Section ~\ref{sec2} of the supplementary material for more details.

\subsection{Evaluation Metrics}
For the gender classifier, the standard evaluation metrics, namely, classification accuracy, Max-Min ratio (the ratio of the best and worst performing subgroups), and Degree of Bias (DoB) (standard deviation of accuracy) are used for fair comparison with the existing studies~\cite{KrishnanR23,ramachandran2022deep,ZhangLM18,DasDB18} on bias evaluation of gender classifier. As a fair model is supposed to have consistent accuracies across all subgroups, this implies that a fair model would have a max-min accuracy ratio closer to $1$ and a DoB closer to $0$. 

For gender-independent facial attribute classifiers, following the studies in ~\cite{zietlow2022leveling,RamaswamyKR21,chuang2021fair}, we used classification accuracy, True Positive Rate (TPR), Difference of Equal Opportunity (DEO) and Difference in Equalized Odds (DEOdds) where Equal Opportunity (EO) requires a classifier to have equal TPRs on each subgroup and a violation of this equal opportunity is measured by the DEO. DEOdds measures the absolute difference in the probability of correctly predicting the positive class between the subgroups for each actual outcome, summed over all possible outcomes~\cite{zietlow2022leveling,chuang2021fair,RamaswamyKR21}. Furthermore, we also analyzed the maximum group accuracy and the minimum group accuracy associated with the best and worst performing demographic subgroups, respectively.
 
A model that can maintain or improve accuracy and TPR while reducing DEO and DEOdds would be an ideal classifier in terms of enhancing accuracy as well as fairness.
\section{Results}

\subsection{Face-based Gender Classification}
In this section, we will discuss the performance and fairness of the face-based gender classifier across gender-racial groups. \\

\noindent \textbf{Intra-Dataset Evaluation:} Table~\ref{tab:table1} shows the performance of the baseline ResNet50 model and our proposed FineFACE model in gender classification when trained and tested on the FairFace dataset. 
As can be seen, our proposed FineFACE model reduced the Degree of Bias (DoB) and the Max-Min accuracy ratio by approximately $86\%$ and $13\%$, respectively, over the baseline. At the same time, the overall classification accuracy improved by about $3\%$ over the baseline ResNet50 model. Note, we also evaluated and compared performance of the baseline DenseNet architecture over FineFACE using DenseNet backbone for gender classification. The experimental results demonstrated the efficacy of FineFace in improving accuracy and fairness over the baseline DenseNet architecture. Thus, highlighting the importance of systematic construction of experts from shallow to deep layers followed by attention region prediction and multi-stage learning by FineFACE over feature reuse between shallow to deep layers by the baseline DenseNet. More details on the experimental results are given in Section ~\ref{subsec1} of the supplementary material. \\

\begin{figure}
    \centering
    \includegraphics[width=1\linewidth]{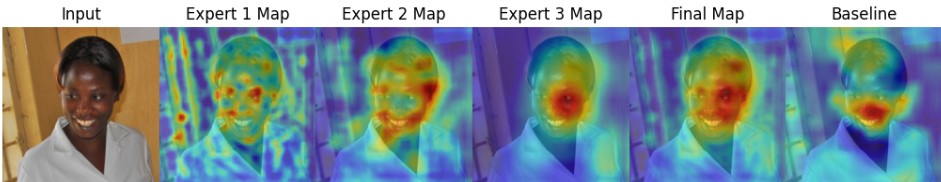}
    \includegraphics[width=1\linewidth]{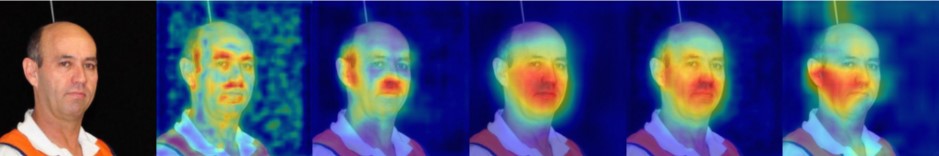}
    \caption{Visualization Results of Gender Classifier. Left through right in each set of images are the input image from FairFace dataset, visualization results based on our FineFACE method's 3 experts ($\tilde{\Omega}_{1}^{norm}$, $\tilde{\Omega}_{2}^{norm}$, $\tilde{\Omega}_{3}^{norm}$), and our method's final visualization ($\tilde{\Omega}_{oval}^{norm}$), versus a basic ResNet50 architecture's final visualization ($\tilde{\Omega}_{ori}^{norm}$). Our FineFACE captures a more comprehensive feature representation of the image, thereby enhancing fairness as well as accuracy.}
    \label{fig:my_cam_viz2}

\end{figure}

\noindent \textbf{Cross-Dataset Evaluation:}
Table~\ref{tab:tableutklfw} shows the results of the baseline ResNet$50$ and our proposed FineFACE, based on ResNet$50$ backbone when trained on FairFace and evaluated on UTKFace and LFWA+ datasets. Our model significantly reduced the bias of the gender classifier by reducing DOB by approximately $55\%$ and $77\%$, and Max-Min ratio by $18\%$ and $17\%$ over the baseline even on the cross-dataset evaluation, respectively. Overall, performance degradation of the classifiers is minimal on cross-dataset evaluation except for the UTKFace dataset due to poor quality samples majorly showing age progression. Table~\ref{tab:tableceleba} shows the results on the CelebA test set. Our model reduced DOB by approximately $41\%$ and Max-Min ratio by $2\%$. 
These results demonstrate the efficacy of our model in significantly reducing bias as well as improving accuracy even on the cross-dataset evaluation.

Fig.~\ref{fig:my_cam_viz2} shows the visualization of the attention map learned by the ResNet50-based FineFACE and the baseline ResNet50-based gender classifier. 
For proposed FineFACE, we generate $4$ heatmaps for each image, i.e., $\tilde{\Omega}_{1}^{norm}$ from expert $1$, $\tilde{\Omega}_{2}^{norm}$ from expert $2$, $\tilde{\Omega}_{3}^{norm}$ from expert $3$ and $\tilde{\Omega}_{oval}^{norm}$ which is the aggregation of the three experts' heatmaps and is used for the final prediction (refer to Section~\ref{subsecarp}). The maps generated by Expert $1$ show a focus on low-level features such as edges, evident from the scattered attention across the face, capturing details such as the outline of the face, eyes, nose, and mouth. The maps generated by Expert $3$ have more concentrated attention on key facial regions that are critical for gender classification, such as the central face area. Thus, there is a clear progression in the focus of attention from Expert $1$ to Expert $3$ and all the varying levels of attention are captured in the final concatenated map (Final Map).
As the original ResNet50 has only a $1$ classifier for prediction, we generated $1$ heatmap $\tilde{\Omega}_{ori}^{norm}$ using the feature maps from the last convolutional layer. 
Note, $\tilde{\Omega}_{3}^{norm}$ and $\tilde{\Omega}_{ori}^{norm}$ are both generated based on the feature maps of the last convolutional layer of the ResNet$50$ backbone, but $\tilde{\Omega}_{3}^{norm}$ captures much more comprehensive and accurate information than $\tilde{\Omega}_{ori}^{norm}$. Further, the overall feature map from the proposed FineFACE model illustrates the efficacy of the fine-grained framework in capturing comprehensive and discriminant regions vital for gender classification over the baseline.

\begin{table}[ht]
\centering
\resizebox{\textwidth}{!}{
\begin{tabular}{cccccccccccccccccc}
\hline
Race     & \multicolumn{2}{c}{White} & \multicolumn{2}{c}{Black} & \multicolumn{2}{c}{East Asian} & \multicolumn{2}{c}{SE Asian} & \multicolumn{2}{c}{Latino} & \multicolumn{2}{c}{Indian} & \multicolumn{2}{c}{Middle E} & \multicolumn{3}{c}{} \\ \hline
Gender   & M & F & M & F & M & F & M & F & M & F & M & F & M & F & \begin{tabular}[c]{@{}c@{}}Max/\\ Min↓\end{tabular} & Overall↑ & DoB↓ \\ \hline
Baseline & 96.5 & 89.9 & 94.4 & 82.4 & 97.2 & 88.9 & 94.4 & 91.5 & 95.6 & 92.2 & 98.1 & 93.3 & 97.8 & 92.4 & 1.18 & 93.2 & 4.2 \\ \hline
FineFACE & 97.1 & 97 & 97.2 & 96.2 & 97 & 96.2 & 96.2 & 97 & 96 & 96.3 & 96.6 & 95.9 & 96.1 & 95.1 & \textbf{1.02} & \textbf{96.4} & \textbf{0.6} \\ \hline
\end{tabular}
}
\caption{Gender Classification Accuracy (\%) on FairFace testset across different demographics using baseline ResNet50 and our proposed FineFACE. M stands for male and F stands for female. The top performance results are highlighted in bold.}
\label{tab:table1}
\vspace{-15pt}
\end{table}

\begin{table}[]
\centering
\resizebox{\textwidth}{!}{%
\begin{tabular}{ccccccccccccccc}
\hline
                         & Race     & \multicolumn{2}{c}{White} & \multicolumn{2}{c}{Black} & \multicolumn{2}{c}{Asian} & \multicolumn{2}{c}{Indian} & \multicolumn{2}{c}{\begin{tabular}[c]{@{}c@{}}Others/\\ Undefined\end{tabular}} &               &               &              \\ \hline
Dataset                  & Gender   & M           & F           & M           & F           & M           & F           & M            & F           & M                                      & F                                      & Max/Min↓       & Overall↑       & DoB↓          \\ \hline
\multirow{2}{*}{UTKFace} & Baseline & 90.2        & 72.2        & 94.6        & 67.6        & 88.2        & 65.7        & 95.1         & 74.9        & 89.1                                   & 78.8                                   & 1.45          & 81.9          & 11.1         \\ \cline{2-15} 
                         & FineFACE & 91          & 86.5        & 93.9        & 79.7        & 91.1        & 81.1        & 95.1         & 86.3        & 90                                     & 88.3                                   & \textbf{1.19} & \textbf{88.5} & \textbf{5}   \\ \hline
\multirow{2}{*}{LFWA+}   & Baseline & 96.9        & 89.1        & 98.7        & 78.8        & 96.5        & 78.3        & 97.9         & 95          & 96.8                                   & 91.1                                   & 1.26          & 95.3          & 7.7          \\ \cline{2-15} 
                         & FineFACE & 99.1        & 98          & 98.9        & 95.3        & 98.6        & 94.8        & 100          & 100         & 98.4                                   & 96.2                                   & \textbf{1.05} & \textbf{98.6} & \textbf{1.9} \\ \hline
\end{tabular}%
}
\caption{Cross-dataset evaluation - Gender Classification Accuracy (\%) on UTKFace and LFWA+ test sets across different demographics for baseline ResNet50 and our proposed FineFACE.}
\label{tab:tableutklfw}
\vspace{-22pt}
\end{table}

\begin{table}[]
\centering
\begin{tabular}{cccccc}
\hline
Gender   & M    & F    & Max/Min↓ & Overall↑ & DoB↓ \\ \hline
Baseline & 90.6 & 94.9 & 1.05    & 93.2    & 2.2 \\ \hline
FineFACE & 96.4 & 99   & \textbf{1.03} & \textbf{98} & \textbf{1.3} \\ \hline
\end{tabular}
\caption{Cross-dataset evaluation - Gender Classification Accuracy (\%) on CelebA testset across gender for baseline Resnet50 and our proposed FineFACE.}
\label{tab:tableceleba}
\vspace{-15pt}
\end{table}

\noindent \textbf{Comparison with Published Work:} Table \ref{tab:tableexp1comp} shows the performance of our proposed FineFACE method over published bias mitigation techniques based on multi-tasking~\cite{DasDB18}, adversarial debiasing ~\cite{ZhangLM18}, deep generative views~\cite{ramachandran2022deep}, and consistency regularization~\cite{KrishnanR23}. All these studies are reported for the ResNet$50$ based gender classifier trained and tested on the FairFace dataset.
As can be seen,  
our proposed FineFACE obtained the lowest DoB of $0.26$ and the Max-Min accuracy ratio of $1.008$ over all the existing published studies. Moreover, the overall accuracy was not only maintained but also increased by $1.74\%$ compared to the second-best model (indicated as D in the Table) based on Deep generative views~\cite{ramachandran2022deep}. Therefore, our proposed method obtains state-of-the-art performance.

Further, existing bias mitigation techniques based on adversarial debiasing~\cite{ZhangLM18} and multitasking~\cite{DasDB18} need demographically annotated data during training. The generative techniques based on deep generative views~\cite{ramachandran2022deep} and consistency regularization ~\cite{KrishnanR23} are computationally very expensive and obtain low generalizability. Compared to the existing methods, the proposed FineFACE offers significant advantages: it mitigates bias in the absence of protected attributes, offers high generalizability, and is application-agnostic. Importantly, our method significantly improves fairness along with overall classification accuracy, emphasizing the importance of fine-grained classification.

\begin{table}[]
\resizebox{\textwidth}{!}{%
\begin{tabular}{ccccccccccc}
\hline
Method & \multicolumn{8}{c}{Accuracy}                                                                                                                                                                                                                                                      & DoB↓           & Max/Min↓        \\ \hline
       & Black & \begin{tabular}[c]{@{}c@{}}East\\ Asian\end{tabular} & Indian & \begin{tabular}[c]{@{}c@{}}Latino\\ Hispanic\end{tabular} & \begin{tabular}[c]{@{}c@{}}Middle\\ Eastern\end{tabular} & \begin{tabular}[c]{@{}c@{}}Southeast\\ Asian\end{tabular} & White & Overall↑        &               &                \\ \hline
A      & 91.26 & 94.45                                                & 95.05  & 95.19                                                     & 97.35                                                    & 94.2                                                      & 94.96 & 94.64          & 1.81          & 1.067          \\ \hline
B      & 87.66 & 91.93                                                & 93.67  & 93.8                                                      & 95.96                                                    & 91.81                                                     & 93.96 & 92.69          & 2.62          & 1.095          \\ \hline
C      & 90.83 & 93.6                                                 & 94.48  & 94.7                                                      & 95.94                                                    & 93.64                                                     & 94.57 & 94             & 1.59          & 1.056          \\ \hline
D      & 91.64 & 95.29                                                & 95.38  & 95.32                                                     & 97.11                                                    & 93.5                                                      & 94.92 & 94.72          & 1.72          & 1.06           \\ \hline
FineFACE   & 96.21 & 96.84                                                & 96.37  & 96.61                                                     & 96.53                                                    & 96.04                                                     & 96.55 & \textbf{96.46} & \textbf{0.26} & \textbf{1.008} \\ \hline
\end{tabular}
}
\caption{Comparative Analysis with FineFACE. A: Multi-Tasking ~\cite{DasDB18} , B: Adversarial debiasing ~\cite{ZhangLM18}, C: Consistency Regularization ~\cite{KrishnanR23} D: Deep Generative Views based  ~\cite{ramachandran2022deep}. The top performance results are highlighted in bold.}
\label{tab:tableexp1comp}
\vspace{-23pt}
\end{table}

\subsection{Gender-independent Facial Attribute Classification}
In this section, we will discuss the performance of our $13$ gender-independent facial attribute classifiers. We report mean scores over the 13 labels~\cite{RamaswamyKR21} called gender-independent target attribute (refer to Section~\ref{secexpdeets} for more details on the $13$ target attributes) with gender as the protected attribute.
Table~\ref{tab:table13labels} shows the performance of the gender-independent facial attribute classifier using our proposed FineFACE over the baseline ResNet50 model. FineFACE improves overall accuracy, minimum group accuracy, and TPR, while significantly reducing bias by approximately $91\%$ (DEO) and $92\%$ (DEODD). There is a marginal reduction in maximum group accuracy by only $1.52\%$.
Worth mentioning, facial attributes like "bags under eyes”, “chubby”,  “high cheekbones”, “narrow eyes”, and “smiling” are more subtle in nature. Thus more detailed features from lower layers help in understanding the subtle cues differentiating a normal cheekbone from a high cheekbone, a narrow eye from a normal eye, and a natural curvature of lips versus a smile across gender. Thus, obtaining performance enhancement as well as bias reduction for these gender-independent facial attributes.

\begin{table}[]
\centering
\resizebox{\textwidth}{!}{%
\begin{tabular}{ccccccccc}
\hline
Method   & Accuracy↑       & \begin{tabular}[c]{@{}c@{}}Max. grp.\\ Acc.\end{tabular} & \begin{tabular}[c]{@{}c@{}}Min. grp.\\ Acc.\end{tabular} & TPR↑            & \begin{tabular}[c]{@{}c@{}}Max. grp.\\ TPR\end{tabular} & \begin{tabular}[c]{@{}c@{}}Min. grp.\\ TPR\end{tabular} & DEO↓           & DEODD↓         \\ \hline
Baseline & 92.47          & \textbf{94.46}                                           & 90.14                                                    & 67.9           & 73.88                                                   & 61.34                                                   & 12.54         & 16.54         \\ \hline
FineFACE & \textbf{92.85} & 92.94                                                    & \textbf{92.76}                                           & \textbf{76.48} & \textbf{76.97}                                          & \textbf{75.79}                                          & \textbf{1.18} & \textbf{1.38} \\ \hline
\end{tabular}
}
\caption{Facial Attribute Classification Accuracy (\%) on the CelebA dataset for baseline and our proposed FineFACE - mean scores over the 13 attributes~\cite{RamaswamyKR21} called gender-independent are reported. The top performance results are highlighted in bold.}
\label{tab:table13labels}
\vspace{-20pt}
\end{table}

\noindent \textbf{Comparison with Published Work:} In this section, we compare the performance of our proposed FineFACE over bias mitigation techniques namely, domain-independent models~\cite{WangQKGNHR20} (Domain Indep.), regularization~\cite{padala2020fnnc,WickpT19} (Regularized), FairMixup ~\cite{chuang2021fair}, GAN-based offline dataset debiasing~\cite{RamaswamyKR21} (GAN Debiasing), and adaptive sampling~\cite{zietlow2022leveling} (g-SMOTE + Adaptive Sampling), reported for the $13$ gender-independent facial attribute classifiers.

\begin{table}[]
\resizebox{\textwidth}{!}{%
\begin{tabular}{lcccccccccc}
\hline
Method         & \begin{tabular}[c]{@{}c@{}}Weight\\-ing*\end{tabular} & \begin{tabular}[c]{@{}c@{}}Domain\\ Indep.*\end{tabular} & \begin{tabular}[c]{@{}c@{}}Baseline\\ single task\end{tabular} & \begin{tabular}[c]{@{}c@{}}GAN\\ Debiasing\end{tabular} & \begin{tabular}[c]{@{}c@{}}Regular\\-ized\end{tabular} & \begin{tabular}[c]{@{}c@{}}g-SMOTE +\\ Adap. Sampl.\end{tabular} & \begin{tabular}[c]{@{}c@{}}g-\\SMOTE\end{tabular} & \begin{tabular}[c]{@{}c@{}}Baseline\\ FairMixup\end{tabular} & \begin{tabular}[c]{@{}c@{}}Fair\\Mixup\end{tabular} & \begin{tabular}[c]{@{}c@{}}FineFACE\\ {[}ours{]}\end{tabular} \\ \hline
Accuracy↑       & 91.45                                                      & 91.24                                                          & 92.47                                                                   & 92.12                                                          & 91.05                                                         & 92.56                                                                          & 92.64                                                    & 92.74                                                                 & 88.46                                                      & \textbf{92.85}                                            \\ 
Max. grp. Acc. & 93.35                                                      & 93.04                                                          & 94.46                                                                   & 94.03                                                          & 94.42                                                         & 94.44                                                                          & \textbf{94.59}                                           & 93.85                                                                 & 90.42                                                      & 92.94                                                     \\
Min. grp. Acc  & 89.06                                                      & 88.93                                                          & 88.93                                                                   & 89.85                                                          & 87.86                                                         & 90.36                                                                          & 90.35                                                    & 91.44                                                                 & 86.36                                                      & \textbf{92.76}                                            \\ \hline
TPR↑            & 64.02                                                      & 70.74                                                          & 67.90                                                                    & 66.13                                                          & 54.2                                                          & 67.11                                                                          & 66.14                                                    & \textbf{79.13}                                                        & 46.67                                                      & 76.48                                                     \\
Max. grp. TPR  & 67.41                                                      & 75.61                                                          & 73.88                                                                   & 70.36                                                          & 56.11                                                         & 74.06                                                                          & 73.43                                                    & \textbf{80.89}                                                        & 47.85                                                      & 76.97                                                     \\
Min. grp. TPR  & 59.74                                                      & 66.05                                                          & 61.34                                                                   & 61.25                                                          & 52.34                                                         & 59.78                                                                          & 58.32                                                    & 72.92                                                                 & 44.27                                                      & \textbf{75.79}                                            \\ \hline
DEO↓            & 7.67                                                       & 9.56                                                           & 12.54                                                                   & 9.11                                                           & 3.77                                                          & 14.28                                                                          & 15.11                                                    & 7.97                                                                  & 3.58                                                       & \textbf{1.18}                                             \\
DEODD↓          & 9                                                          & 13.29                                                          & 16.54                                                                   & 12.04                                                          & 5.06                                                          & 19.3                                                                           & 19.32                                                    & 10.06                                                                 & 4.29                                                       & \textbf{1.38}                                             \\ \hline
\end{tabular}%
}
\caption{Fairness methods on the CelebA dataset - mean scores over the 13 labels~\cite{RamaswamyKR21} called gender independent. The top performance results are highlighted in bold.}
\label{tab:13_attributes_comparison}
\end{table}

We summarized the results in the Table~\ref{tab:13_attributes_comparison}. 
The proposed FineFACE has achieved improved performance in both overall accuracy and accuracy of the worst-performing group compared to the baseline classifier. Although there is a slight reduction in the accuracy of the best-performing group (by $1.52\%$), the performance of this group remains comparable to the overall and worst-performing groups which improve by approximately $0.4\%$ and $2.5\%$ respectively. Worth mentioning, among $6$ existing bias mitigation techniques in Table~\ref{tab:13_attributes_comparison}, only g-SMOTE~\cite{RamaswamyKR21} and g-SMOTE adaptive sampling~\cite{zietlow2022leveling} are able to improve performance of the worst performing group, with respect to the baseline, along with our proposed FineFACE, thus obtaining~\textbf{pareto-efficiency}. Furthermore, our method obtains the highest improvement in the TPR of the worst-performing groups, along with the second-best results for overall TPR and the TPR for the best-performing groups. The \textbf{key highlight} of our method is the substantial reduction in both DEO and DEOdds, $3\times$ lower than the next best method i.e., Fair Mixup~\cite{chuang2021fair}. Comparison of the various fairness methods is visually represented in Fig. ~\ref{fig:methods_comparison2}, Section ~\ref{subsec2} of the supplementary material

We also reported the minimum group accuracy of the baseline ResNet50 model and our FineFACE for each of the $13$ gender-independent attributes individually in Table ~\ref{tab:min_grp_acc_13_attrs}. Our model outperformed the baseline for all except $1$ attributes ("Narrow Eyes" by the baseline is more accurate by approximately $4\%$). For the other $12$ attributes, our model outperformed at least by $0.5\%$ and
at most by $7.9\%$. Thus, we also demonstrate the efficacy of our FineFACE in improving the minimum group accuracy for the majority ($12$ out of $13$ attributes) of facial attributes on an individual basis.

\begin{table}[]
\centering
\begin{tabular}{l|ll}
\hline
\textbf{Attribute Name} & \textbf{Baseline} & \textbf{FineFACE} \\ \hline
Bags Under Eyes         & 73.24             & 80.73             \\
Bangs                   & 94.67             & 95.9              \\
Black Hair              & 86.4              & 90.51             \\
Blond Hair              & 91.96             & 94.41             \\
Brown Hair              & 81.26             & 89.14             \\
Chubby                  & 89.29             & 95.64             \\
Eyeglasses              & 99.23             & 99.73             \\
Gray Hair               & 95.28             & 98.35             \\
High Cheekbones         & 85.53             & 87.83             \\
Mouth Slightly Open     & 93.46             & 94.23             \\
Narrow Eyes             & 91.97             & 87.99             \\
Smiling                 & 91.64             & 93.17             \\
WearingHat              & 98.21             & 99.08             \\ \hline
\end{tabular}
\caption{Minimum Group Accuracy for the $13$ gender-independent individual attributes.}
\label{tab:min_grp_acc_13_attrs}
\end{table}
\vspace{-15pt}
\section{Conclusion}
The task of facial attribute classification presents inherent complexities due to high inter-class similarity, significant intra-class variation, and demographic diversity, which often result in performance disparities across protected attributes.

To effectively tackle these challenges, it is essential to incorporate local and subtle cues into the classification process. In our research, we propose a novel fine-grained feature framework designed for demographically fair facial attribute classification. This framework integrates detailed low-level and semantic high-level information across shallow to deep layers of the model. 
Through extensive evaluation on widely used facial attribute datasets, our approach demonstrates significant effectiveness in learning fair representation, achieving up to a three-fold reduction in bias compared to state-of-the-art bias mitigation techniques. Importantly, our method achieves a Pareto-efficient balance between accuracy and fairness without requiring the presence of protected attribute labels during classifier training—a critical advantage given privacy concerns and regulatory constraints that often prohibit the collection of such sensitive data. Furthermore, our study marks the first benchmark evaluation of the fairness of facial attribute classifiers using fine-grained features compared to existing supervised bias mitigation techniques.  

While the multi-step training strategy extends the training duration compared to the original backbone networks, the training is an offline process and the more significant concern in real-world applications is the inference cost which is affordable for our method. As part of future work, we will also explore other backbone architectures such as Transformer.
In addition, we will further analyze biases across intersectional groups, such as {gender + target attribute}, following insights from recent studies~\cite{zietlow2022leveling,chuang2021fair}. \\ 
\noindent\textbf{Acknowledgements} This work is supported by National Science Foundation (NSF) award no. $2129173$.

 \bibliographystyle{splncs04}
 \bibliography{references,refs,ref}

\appendix

\section*{Supplementary Material}

This supplementary material includes the following:

\begin{itemize}
    \item Additional Experimental Results (Section \ref{sec1})
    \begin{itemize}
        \item Gender Classification using DenseNet Backbone (Subsection \ref{subsec1})
        \item 13 Gender-independent Facial Attribute Classification (Subsection \ref{subsec2})
    \end{itemize}
    \item Ablation Studies (Section \ref{sec2})
    \item Pseudo-code of the Training Algorithm (Section \ref{sec3})
\end{itemize}

\section{Additional Experimental Results}\label{sec1}

\subsection{Gender Classification using DenseNet Backbone}\label{subsec1}
We also investigated the potential of our proposed FineFACE for gender classification using the DenseNet backbone. Unlike traditional networks where each layer feeds into the next layer, DenseNet connects each layer to every subsequent layer in a feedforward fashion. This means that the output feature maps of all previous layers are concatenated together as inputs to the subsequent layers. As the dense connectivity of this architecture ensures maximum information flow between layers and encourages feature reuse, it would be interesting to evaluate the fairness of the DenseNet-based gender classifier over our proposed FineFACE based on the DenseNet backbone. 

For the baseline, we used the DenseNet-$121$ variant. For FineFACE, we developed three experts - e1 covered the dense blocks $1$ and $2$, e2 covered the dense blocks $1$, $2$, and $3$, and e3 covered all the $4$ dense blocks of DenseNet-$121$. The model was trained with the same progressive multistep strategy discussed in Section $3$ of the main paper. Initially, each expert focused on cues from its own blocks. In the final steps, all experts were trained together to consolidate their predictions and attention regions. The model is trained using the same hyperparameters mentioned in Section $4.2$ of the main paper.

\begin{table}[]
\centering
\resizebox{\textwidth}{!}{
\begin{tabular}{cccccccccccccccccc}
\hline
Race     & \multicolumn{2}{c}{White} & \multicolumn{2}{c}{Black} & \multicolumn{2}{c}{East Asian} & \multicolumn{2}{c}{SE Asian} & \multicolumn{2}{c}{Latino} & \multicolumn{2}{c}{Indian} & \multicolumn{2}{c}{Middle E} &               &               &              \\ \hline
Gender   & M           & F           & M           & F           & M              & F             & M             & F            & M            & F           & M            & F           & M             & F            & Max/Min       & Overall       & DOB          \\ \hline
DenseNet & 94.5        & 94.1        & 95          & 95.8        & 94.1           & 95.9          & 95            & 94.3         & 94.6         & 94.6        & 94.2         & 97          & 93.9          & 95.9         & 1.03          & 94.8          & 0.9          \\ \hline
FineFACE & 97.3        & 97.1        & 97.1        & 96.4        & 97.2           & 96.1          & 96.4          & 97           & 96.1         & 96.2        & 96.7         & 95.9        & 96.2          & 95.5         & \textbf{0.98} & \textbf{96.5} & \textbf{0.6} \\ \hline
\end{tabular}
}
\caption{Gender Classification Accuracy (\%) on FairFace test set – FineFACE vs. DenseNet. M stands for Male, and F stands for Female. The top performance results are highlighted in bold.}
\label{tab:suppltable1}
\end{table}

Table~\ref{tab:suppltable1} shows the performance of the DenseNet-based architecture for gender classification when trained and tested on the FairFace dataset. As can be seen, the DenseNet-based gender classifier performs consistently across demographics with high overall accuracy as well as low DOB over the ResNet50 model (refer to Table 2, Section 5.1 of the main paper) by $1.6\%$ increment and $78.5\%$ decrement, respectively. However, FineFACE with the DenseNet backbone still outperforms the baseline DenseNet with an overall accuracy increment of $1.7\%$, and the max-min accuracy ratio and DoB reduction by approximately $4.8\%$ and $33\%$, respectively.

This highlights the importance of systematic construction of experts from shallow to deep layers followed by attention region prediction and multi-stage learning by FineFACE over feature reuse between shallow to deep layers by the baseline DenseNet.

\subsection{13 Gender-independent Facial Attribute Classification}\label{subsec2}

\begin{figure} []
    \centering
    \includegraphics[width=1\linewidth]{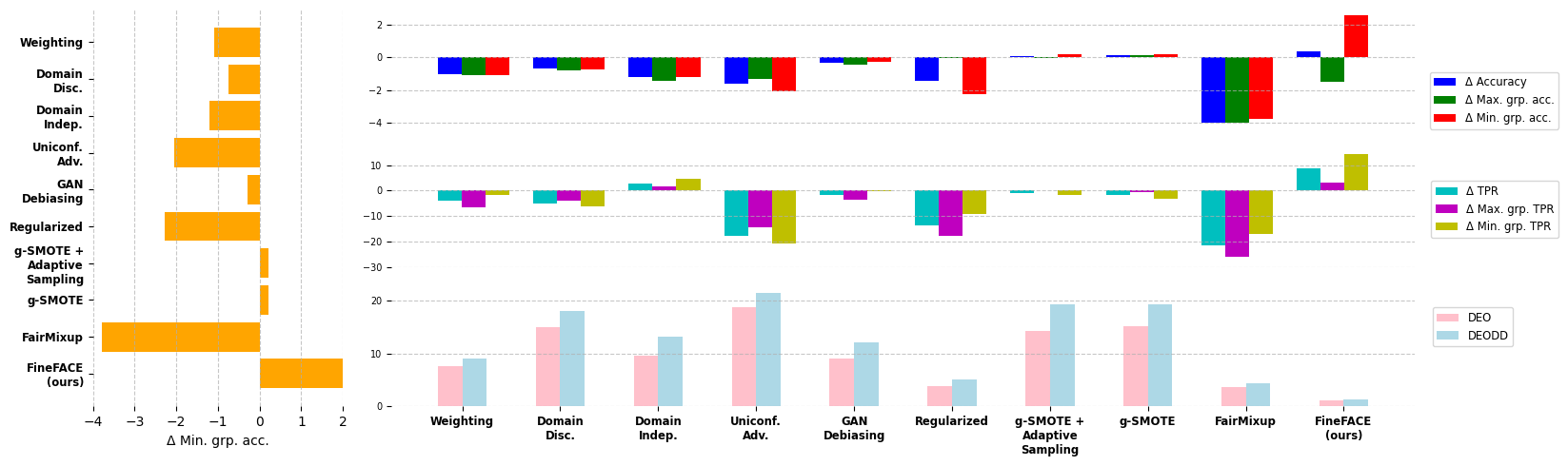}
    \caption{Fairness methods on the CelebA dataset. We report mean scores over 13 gender independent target attributes. Left: Change in minimum group accuracy difference relative to the single task baseline classifier. Top Right: A decomposition of the change in accuracy from baseline into overall change, change in the best performing group, and the worst. Center Right: Decomposition of the change in True Positive Rate (TPR). Bottom Right: plots of DEO and DEOdds.}
    \label{fig:methods_comparison2}
\end{figure}

We also visualized the results of FineFACE for 13 gender-independent attribute classification over SOTA bias mitigation techniques in Fig. ~\ref{fig:methods_comparison2} (refer to Table 7, Section 5.2 of the main paper for further details) with reference to baseline ResNet50 model. Our method is the best in terms of improving the accuracy of the worst performing group while also increasing the overall accuracy. There is a slight reduction in the accuracy of the best performing group while still remaining comparable with the other groups. Only two fairness methods significantly increase the TPR, and our method performs the best with improvements of approx. $8.5\%$, $3\%$, $14.4\%$ on overall, best and worst groups, respectively. Finally, our method obtains the lowest DEO and DEOdds ($1.18$ and $1.38$, respectively).

\section{Ablation Studies}\label{sec2}

To comprehensively analyze our method, we did an ablation studies of the design choices. This subsection chooses the face-based gender classification trained and tested on the FairFace dataset.\\

\noindent \textbf{\textit{Evaluation of Different Pooling Methods for Building Experts.}} As mentioned in Section 3 of the main paper, each expert uses a global max pooling (GMP) layer to aggregate intermediate feature maps into a descriptor. Here we evaluate the effectiveness of this design choice. We used global average pooling (GAP), bilinear pooling (BP) ~\cite{lin2017bilinear}, and global covariance pooling (GCP) ~\cite{wang2020deep} to replace GMP, respectively, and observe the change in accuracy. The accuracy obtained with GMP is $1\%$–$1.4\%$ higher compared to other pooling methods. This improvement is likely because GMP captures sharp features, like edges and corners, and is invariant to various affine transformations, aiding the model in leveraging attentional information from extensive data augmentation. Specifically, for the proposed FineFACE, different experts focus on different attention regions where key objects may be located in various regional positions, and GMP effectively handles this variability. Additionally, GMP reduces the number of parameters, which helps prevent the model from falling into local optima for complex learning tasks due to difficulties in parameter optimization. \\

\noindent \textbf{\textit{Contribution of Fusing the Prediction Scores.}} For the testing procedure, the final prediction score is obtained by fusing the prediction scores given by different experts based on different types of inputs (i.e., raw input and the overall attention region $A_{oval}$). It is necessary to verify the effectiveness of this inference strategy. When the raw image is the input, the average of the prediction scores obtained $0.5\%$–$3.2\%$ higher accuracy than each of the prediction scores separately. Similarly, when $A_{oval}$ is the input, the average of the prediction scores obtains $0.5\%$–$3.6\%$ higher accuracy than each of the prediction scores separately. Furthermore, the fusion of the prediction scores obtained with the raw image and $A_{oval}$ is $0.2\%$–$0.4\%$ higher than using each of them individually. These results verify the complementary nature of the information learned by different experts based on different types of inputs. For all the experiments, we report the accuracy obtained by fusing all the prediction scores.

\section{Pseudo-code of the Training Algorithm}\label{sec3}
In this section, we discuss the pseudo-code of the multi-step mutual learning algorithm detailed in Section $3.3$ of the main paper.

\begin{minipage}{\columnwidth}
\begin{algorithm}[H]
\caption{Multi-step Mutual Learning Algorithm}
\label{alg:ad}
\begin{algorithmic}
\REQUIRE $D = \{(\text{input}^{i}, \text{target}^{i})\}_{i=1}^{l},\ N,\ L_{\text{cls}}(\cdot)$
\FOR{epoch = $1$ to no\_of\_epochs}
\FOR{(input, target) in $D$}
\STATE
\STATE \(\triangleright\) Train each expert from deep to shallow in multiple steps.
\STATE $x_N, \{A_1, A_2$, ..., $A_n$, ..., $A_N$, $A_{oval}$\} $\gets e_N$(input)
\STATE $v_N \gets F_N$($x_N$)
\STATE $p_{N}$ $\gets$  $f_{N}^{clf}(v_{N})$
\STATE $L_{N}$ $\gets$ $L_{cls}$($p_{N}$, target)
\STATE \textbf{BACKPROP}($L_{N}$)
\FOR{n = $N-1$ downto 1 by -1}
\STATE $\text{input}_\text{n}$ $\gets$ Randomly\_choose\_from(\{input, $A_{1}$, $A_{2}$, ..., $A_{n}$, ..., $A_{N}$\})
\STATE $x_{n}$ $\gets$ $e_{n}$($\text{input}_\text{n}$)
\STATE $v_{n}$ $\gets$ $F_{n}$($x_{n}$)
\STATE $p_{n}$ $\gets$  $f_{n}^{clf}(v_{n})$
\STATE $L_{n}$ $\gets$ $L_{cls}$($p_{n}$, target)
\STATE \textbf{BACKPROP}($L_{n}$)
\ENDFOR
\STATE
\STATE \(\triangleright\) Train the experts and concatenation of experts with $A_{oval}$ in one pass.
\FOR{n = $1$ to N}
\STATE $x_{n}^{A}$ $\gets$ $e_{n}$($A_{oval}$)
\STATE $v_{n}^{A}$ $\gets$ $F_{n}$($x_{n}^{A}$)
\STATE $p_{n}^{A}$ $\gets$ $f_{n}^{clf}(v_{n}^{A})$
\ENDFOR
\STATE $v_{oval}^{A}$ $\gets$ $f_{concat}$($v_{1}^{A}$, $v_{2}^{A}$, ..., $v_{N}^{A}$)
\STATE $p_{oval}^{A}$ $\gets$ $f_{oval}^{clf}(v_{oval}^{A})$
\STATE $L^{A}$ $\gets$ $L_{cls}$($p_{1}^{A}$, target) $+$ $L_{cls}$($p_{2}^{A}$, target) $+$....$+$  $L_{cls}$($p_{N}^{A}$, target) $+$ $L_{cls}$($p_{oval}^{A}$, target)
\STATE \textbf{BACKPROP}($L^{A}$)
\STATE
\STATE \(\triangleright\) Train the concatenation of experts with the raw input.
\FOR{n = $1$ to N}
\STATE $x_{n}$ $\gets$ $e_{n}$(input)
\STATE $v_{n}$ $\gets$ $F_{n}$($x_{n}$)
\ENDFOR
\STATE $v_{oval}$ $\gets$ $f_{concat}$($v_{1}$, $v_{2}$, ..., $v_{N}$)
\STATE $p_{oval}$ $\gets$ $f_{oval}^{clf}(v_{oval})$
\STATE $L_{oval}$ $\gets$ $L_{cls}$($p_{oval}$, target)
\STATE \textbf{BACKPROP}($L_{oval}$)
\ENDFOR
\ENDFOR
\end{algorithmic}
\end{algorithm}
\end{minipage}

wherein, l stands for the total number of batches in dataset D, and N represents the number of experts. $L_{cls}$(.) is the cross-entropy loss of the classification task. 

Each iteration of the training takes place in N + 2 steps. In the initial N steps, each expert is trained from deep to shallow, starting with the deepest expert ($e_{N}$ to $e_{1}$). In step $N+1$, all experts and their concatenation are trained with the general attention region $A_{oval}$ in one go. In step $N+2$, the concatenation of all experts is trained with the raw input to ensure that the parameters of $f_{oval}^{clf}(.)$  fit the resolution of the objects in the original input.

\end{document}